\documentclass[letterpaper, 10 pt, conference]{ieeeconf}  %

\usepackage{graphicx}
\usepackage{amsmath}
\usepackage{amssymb}
\usepackage{booktabs}
\usepackage{epsfig}
\usepackage{amsmath}
\usepackage{amssymb}
\usepackage{array}
\usepackage{csquotes}

\usepackage{cancel}

\usepackage{xcolor, colortbl}
\usepackage{tabularx}
\usepackage{multirow}
\let\labelindent\relax
\usepackage{enumitem}
\setlist{leftmargin=5.5mm}
\usepackage{bbm}

\usepackage{pifont}
\usepackage{capt-of}
\usepackage{array,multirow}
\usepackage[export]{adjustbox}

\newcolumntype{L}[1]{>{\raggedright\let\newline\\\arraybackslash\hspace{0pt}}m{#1}}
\newcolumntype{C}[1]{>{\centering\let\newline\\\arraybackslash\hspace{0pt}}m{#1}}
\newcolumntype{R}[1]{>{\raggedleft\let\newline\\\arraybackslash\hspace{0pt}}m{#1}}

\usepackage[breaklinks=true,letterpaper=true,colorlinks,bookmarks=true]{hyperref}

\def\ie{\emph{i.e.}}
\def\eg{\emph{e.g.}}
\def\etal{\emph{et al.}}

\definecolor{brown}{rgb}{0.65, 0.16, 0.16}
\definecolor{purp}{rgb}{0.65, 0.16, 0.65}
\definecolor{orange}{rgb}{0.9, 0.3, 0.1}
\definecolor{Gray}{gray}{0.85}
\definecolor{dblue}{rgb}{0.1, 0.1, 0.75}
\definecolor{green}{rgb}{0, 0.8, 0.2}
\definecolor{grey}{rgb}{0.9, 0.9, 0.9}

\newcommand{\Fig}[1]{Fig.~\ref{fig:#1}}
\newcommand{\Sec}[1]{Sec.~\ref{sec:#1}}
\newcommand{\Eq}[1]{Eq.~(\ref{eq:#1})}
\newcommand{\Tbl}[1]{Table~\ref{tab:#1}}
\newcommand{\ccol}{\cellcolor{grey}}

\newcommand{\cmark}{\ding{51}}%

\newcommand{\expnum}[2]{{#1}\mathrm{e}{-#2}}

\definecolor{lgray}{rgb}{0.96, 0.96, 0.96}
\definecolor{aero}{rgb}{0.5, 0.25, 0.5}
\definecolor{bicy}{rgb}{0.95, 0.14, 0.91}
\definecolor{bird}{rgb}{0.27, 0.27, 0.27}
\definecolor{boat}{rgb}{0.4, 0.4, 0.61}
\definecolor{bottle}{rgb}{0.75, 0.6, 0.6}
\definecolor{bus}{rgb}{0.6, 0.6, 0.6}
\definecolor{car}{rgb}{0.98, 0.66, 0.12}
\definecolor{cat}{rgb}{0.86, 0.86, 0}
\definecolor{chair}{rgb}{0.42, 0.56, 0.14}
\definecolor{cow}{rgb}{0.6, 0.98, 0.6}
\definecolor{table}{rgb}{0.27, 0.51, 0.71}
\definecolor{dog}{rgb}{0.86, 0.08, 0.16}
\definecolor{horse}{rgb}{1, 0, 0}
\definecolor{mbike}{rgb}{0, 0, 0.56}
\definecolor{person}{rgb}{0, 0, 0.27}
\definecolor{plant}{rgb}{0, 0.16, 0.39}
\definecolor{sheep}{rgb}{0, 0.31, 0.39}
\definecolor{sofa}{rgb}{0, 0, 0.9}
\definecolor{train}{rgb}{0.46, 0.04, 0.125}
\definecolor{tv}{rgb}{0, 0.2, 0.6}
\newcommand{\brt}[1]{{\color{lgray}{#1}}}

\IEEEoverridecommandlockouts                              %
\begin{document}

\title{\LARGE \bf
WEDGE: Web-Image Assisted Domain Generalization \\ for Semantic Segmentation
}

\author{
    Namyup Kim$^{1}$, Taeyoung Son$^{2}$, Jaehyun Pahk$^{1}$, Cuiling Lan$^{3}$, Wenjun Zeng$^{4}$, and Suha Kwak$^{1}$ %
	\thanks{$^{1}$ Namyup Kim, Jaehyun Pahk and Suha Kwak are with POSTECH, Pohang, Republic of Korea. 
	\texttt{\{namyup, jhpahk, suha.kwak\}@postech.ac.kr}
	}
	\thanks{$^{2}$ Taeyoung Son is with NALBI, Seoul, Republic of Korea. \texttt{taeyoung@nalbi.ai}}
	\thanks{$^{3}$ Cuiling Lan is with Microsoft Research Asia, Beijing, China. \texttt{culan@microsoft.com}}
	\thanks{$^{4}$ Wenjun Zeng is with EIT Institute for Advanced Study, Beijing, China. \texttt{wenjunzeng@eias.ac.cn}}
}

\maketitle

\begin{abstract}
Domain generalization for semantic segmentation is highly demanded in real applications, where a trained model is expected to work well in previously unseen domains.
One challenge lies in the lack of data which could cover the diverse distributions of the possible unseen domains for training. 
In this paper, we propose a WEb-image assisted Domain GEneralization (WEDGE) scheme, which is the first to exploit the diversity of web-crawled images for generalizable semantic segmentation.
To explore and exploit the real-world data distributions, we collect web-crawled images which present large diversity in terms of weather conditions, sites, lighting, camera styles, etc. 
We also present a method which injects styles of the web-crawled images into training images on-the-fly during training, which enables the network to experience images of diverse styles with reliable labels for effective training.
Moreover, we use the web-crawled images with their predicted pseudo labels for training to further enhance the capability of the network. 
Extensive experiments demonstrate that our method clearly outperforms existing domain generalization techniques.

\end{abstract}

\section{Introduction}
\label{sec:intro}

Semantic segmentation has played crucial roles in many applications like autonomous vehicle and augmented reality.
Recent advances in this field are mainly attributed to the development of deep neural networks, whose success depends heavily on the availability of a large-scale annotated dataset for training. 
However, building a large training dataset is prohibitively expensive since it demands manual annotation of pixel-level class labels.
To mitigate this problem, synthetic image datasets have been introduced~\cite{gta5,synthia}. 
They provide a large amount of labeled images for training at minimal cost of construction.
Also, they can simulate scenes that are rarely observed in the real world yet must be considered in training (\eg, accidents in autonomous driving scenarios).

\begin{figure*}[!t]
\begin{center}
\includegraphics[width=0.99 \linewidth]{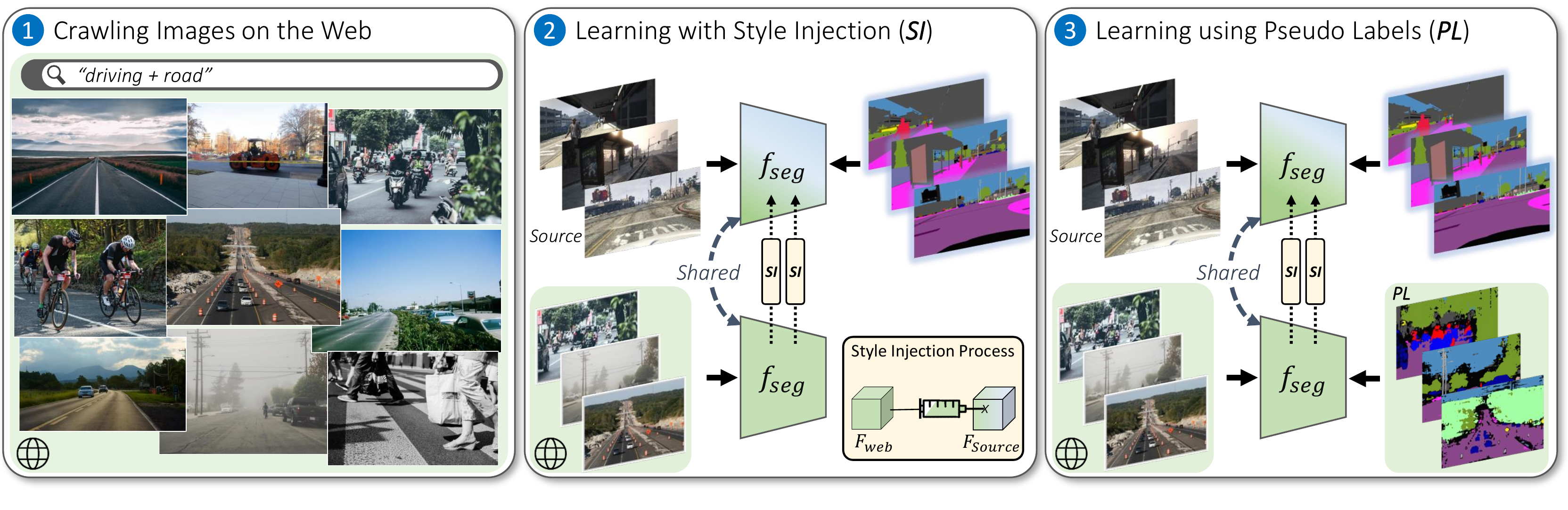}
\end{center}
\vspace{-5mm}
\caption{Overall framework of WEDGE.
(1) Crawling real and task-relevant images from the Web automatically. 
(2) Learning semantic segmentation while transferring feature statistics of web images to features of synthetic training images in the source domain.
(3) Further training the model using both source images and web-crawled images with predicted pseudo labels.
}
\label{fig:overall}
\vspace{-4.5mm}
\end{figure*}

When learning semantic segmentation using synthetic images, it is essential to close the gap between the synthetic and real domains caused by their appearance differences so as to avoid performance degradation of learned models on real-world images.
Most of existing solutions to this issue belong to the category of domain adaptation, which aims at adapting models trained on synthetic images (\ie, source domain) to real-world images (\ie, target domain).
In general, domain adaptation methods assume a single, particular target domain and train models using images from both of labeled source and unlabeled target domains~\cite{hoffman2016fcns,valada2017adapnet,tsai2018learning,li2019bidirectional,zou2019confidence,pan2020unsupervised,NEURIPS2020_243be281,zhang2021target}.
Unfortunately, this setting limits applicability of learned models since, when deployed, models can face multiple and diverse target domains (\eg, geolocations and weather conditions in the case of autonomous vehicle) that are latent at the training stage.

As a more realistic solution to the problem, we study \emph{domain generalization} for semantic segmentation.
The goal of this task is to learn models that generalize well to various target domains without having access to their images in training. 
A pioneer work~\cite{DRPC} achieves the generalization by forcing segmentation models to be invariant to random style variations of input image.
However, this method is costly since it applies an image-to-image translator~\cite{zhu2017unpaired} to every synthetic image multiple times for the style randomization. 
Moreover, random styles are given by a small number of images sampled from ImageNet~\cite{Imagenet}, and thus often irrelevant to target applications and hard to cover a wide range of real-world image styles.
Follow-up research is often limited by the knowledge of ImageNet too. 
For example, Chen~\etal~\cite{chen2020automated} encourage the representations learned using synthetic images to be similar with those of an ImageNet pretrained network, and Huang~\etal~\cite{huang2021fsdr} randomize synthetic images in a frequency space using a small subset of ImageNet as references for stylization.

In this paper, we propose a WEb-image assisted Domain GEneralization scheme, dubbed \emph{WEDGE}, which overcomes the limitations of the previous work by using real and application-relevant images crawled from web repositories (\eg, Flickr).
The crawling process demands no or minimal human intervention as it only asks search keywords that are determined directly by target application (\eg, ``driving + road'' for autonomous driving) or classes appearing in the source domain images. 
Moreover, unlike those of ImageNet, the retrieved images can be used for self-supervised learning as well as for stylization since they are expected to be relevant to target application.

As illustrated in Fig.~\ref{fig:overall},
WEDGE utilizes images crawled from the Web in two different ways.
First, it replaces neural styles of synthetic training images with those of web-crawled images on-the-fly during training.
This helps enhance the generalization by giving illusions of diverse real images while exploiting groundtruth labels of synthetic images.
For this purpose, we introduce a \emph{style injection} module that conducts the style manipulation in a feature level at low cost.
Since it is substantially more efficient than the image-to-image translator used in~\cite{DRPC}, it allows to perform the stylization on-the-fly using a large number of web images as style references in training.

Second, the web-crawled images are used as additional training data with pseudo segmentation labels.
To this end, the entire training procedure is divided into two stages.
In the first stage, a segmentation model is trained with the style injection module, and the web-crawled images are used only for stylization.
The learned model is then applied to the web-crawled images to estimate their pseudo labels.
The second stage is identical to the former, except that it also utilizes the web-crawled images as training data by taking their pseudo labels as supervision.

To demonstrate the efficacy of WEDGE, we adopt each of the GTA5~\cite{gta5} and SYNTHIA~\cite{synthia} datasets as the source domain for training, and evaluate trained models on three different real image datasets~\cite{cityscapes,BDD100k,mapillary}.
Experimental results demonstrate that WEDGE enables segmentation models to generalize well to multiple unseen real domains and clearly outperforms existing methods. 
In summary, the contribution of this paper is three-fold:
\vspace{-0.5mm}
\begin{itemize}[leftmargin=4mm]
    \itemsep=0.1mm
    \item To the best of our knowledge, WEDGE is the first that attempts to utilize web-crawled images for domain generalizable semantic segmentation. %
    These images facilitate self training based on the realistic data which may better approximate unseen testing domains.
    \item We introduce style injection to domain generalizable semantic segmentation.
    Through web-crawled images, it helps achieve the generalization by giving diverse illusions of reality to the network being trained using labeled synthetic images. Also, the superiority of our particular style injection method over other potential candidates is demonstrated empirically.
    \item WEDGE outperforms existing domain generalization techniques~\cite{DRPC,chen2020automated,choi2021robustnet,huang2021fsdr,kim2022pin} in every experiment.
\end{itemize}

\section{Related Work}
\label{sec:relatedwork}

\vspace{0.2mm} \noindent \textbf{Domain generalizable semantic segmentation.}
The goal of domain generalization is to learn models that well generalize to unseen domains~\cite{muandet2013domain,gan2016learning}.
Early approaches address this task mostly for classification~\cite{li2018domain,li2017deeper,li2018learning,pan2018two,nam2019reducing,zhou2021domain}, but recent research demonstrates its potential for semantic segmentation~\cite{DRPC,chen2020automated,pan2018two}.
For example, Pan~\etal~\cite{pan2018two} tackle this problem by feature normalization for learning domain invariant features, and Chen~\etal~\cite{chen2020automated} encourage the representation learned on a source domain to be similar with that of an ImageNet pretrained model.
Also, Yue~\etal~\cite{DRPC} propose to learn features invariant to random style variations of input, and establish 
an evaluation protocol for 
the task.
The main difference of ours from the previous work is that ours explores and exploits real images on the Web which enable models to experience a variety of real domains during training with no human intervention. 

\vspace{0.2mm} \noindent \textbf{Neural style transfer.}
A pioneer work by Gatys~\etal~\cite{gatys2016image} shows that an image style can be captured by the Gram matrix of a feature map, and Johnson~\etal~\cite{johnson2016perceptual} further enhance this idea to transfer a neural style to arbitrary images.
Huang~\etal~\cite{AdaIN} demonstrate that the channel-wise mean and standard deviation of a feature map represent image style effectively.
Also, Nam and Kim~\cite{nam2018batch} and Kim~\etal~\cite{Kim2020UGATIT} propose to use different normalization operations complementary to each other for style transfer.
Recently, content-aware style transfer methods~\cite{park2019arbitrary,liu2021adaattn,huo2021manifold}
are emerged to catch more details of local style patterns and to preserve content better.
Huo~\etal~\cite{huo2021manifold} suppose that features passing through a network form a manifold per each semantic region, and present a new style transfer technique based on manifold alignment.
Style injection in WEDGE is motivated particularly by the techniques presented in~\cite{AdaIN,huo2021manifold}.
However, it is distinct from them in that it aims to perform feature stylization, rather than image stylization.

\vspace{0.2mm} \noindent \textbf{Learning using data on the Web.}
Modern recognition models tend to be data-hungry, yet the amount of training data is usually limited.
Data on the Web have been exploited to alleviate this issue.
Early studies utilize web-crawled images and videos for learning concept recognition by using their search keywords as pseudo labels~\cite{chen2013neil,divvala2014learning,chen2015webly}, and for object localization via clustering images~\cite{chen2013neil,divvala2014learning} or by motion segmentation~\cite{prest2012learning}.
Motivated by recent advances in pseudo labeling,
a large-scale web data have been used for supervised learning with their pseudo labels, which is known as webly supervised learning.
For image classification, 
Niu~\etal~\cite{niu2018webly} present a reliable way of utilizing search keywords as pseudo class labels.
For semantic segmentation, Hong~\etal~\cite{Hong2017_webly} and Lee~\etal~\cite{lee2019frame} compute pseudo labels by segmenting web videos using attentions drawn by an image classifier.
Motivated by these, we present the first that makes use of web images for domain generalization.

\begin{figure}[!t]
\begin{center}
\includegraphics[width=\linewidth]{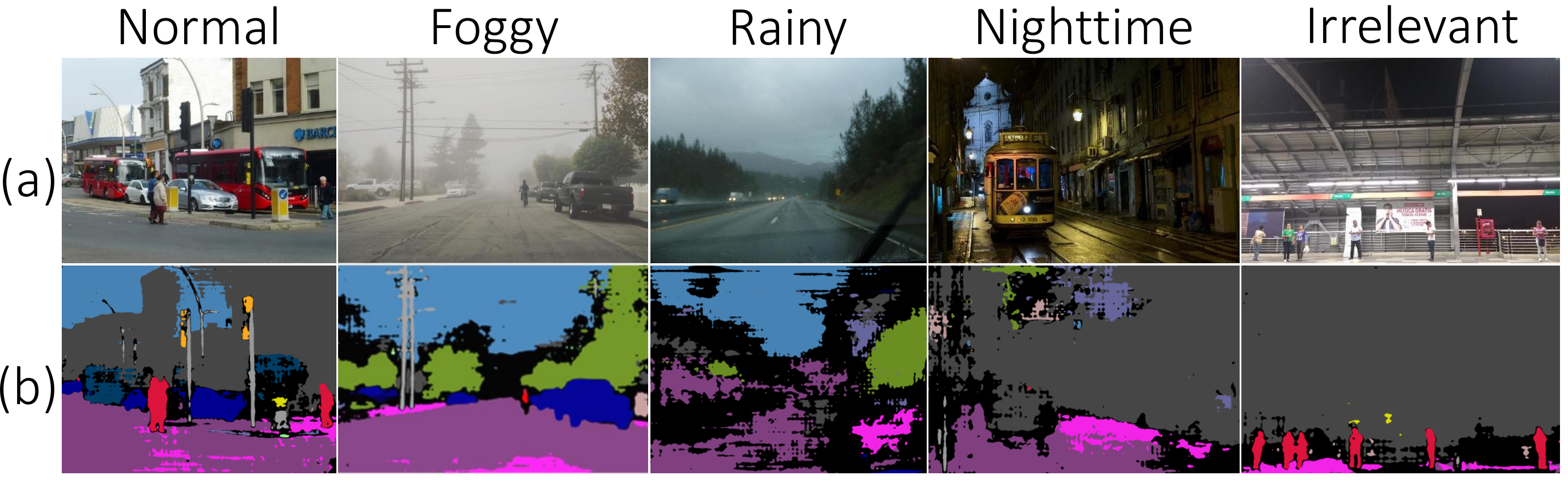}
\end{center}
\vspace{-5mm}
\caption{
Qualitative examples of web images and their pseudo labels generated by the segmentation model with ResNet101 backbone trained by the first stage of WEDGE.
(a) Input images. (b) Pseudo segmentation labels.
These images demonstrate large diversity of the web images, which is vital for achieving generalization to multiple latent test domains.
} 
\label{fig:pseudo_labels_qual}
\vspace{-4mm}
\end{figure}

\section{Proposed Method}
As shown in Fig.~\ref{fig:overall}, WEDGE is divided into three steps: %
\begin{itemize}[leftmargin=5mm] 
    \item[1.] \textbf{Crawling images from web repositories automatically.}
    \item[2.] \textbf{The first stage of training with style injection (SI)}. Learning a segmentation model on the synthetic dataset while injecting styles of the web-crawled images to its intermediate features for training.
    \item[3.] \textbf{The second stage of training using pseudo labels (PL)}. Further training the model using the web-crawled images and their pseudo segmentation labels as well as the synthetic dataset.
\end{itemize}
Details of each step are given in the remainder of this section.

\subsection{Crawling Images from the Web}
\label{sec:crawl_images}
We collect 4,904 images by crawling on Flickr, through the search keyword ``driving + road'' to find images relevant to the target application scenario, \ie, autonomous driving. 
Examples of the collected images are presented in Fig.~\ref{fig:pseudo_labels_qual}.

Using these images for domain generalization has several advantages.
First, they offer a large variety of real image styles as illustrated in~Fig.~\ref{fig:pseudo_labels_qual},
which \emph{potentially cover testing domains}.
This is vital for achieving generalization to unseen domains. 
Second, they are not random but mostly relevant to target applications due to the use of search keywords and thus can be used for supervised learning given their pseudo labels. 
Last, they are accessible with minimal human intervention since the crawling process above is fully automated given a query.
Note that %
previous work~\cite{DRPC, chen2020automated, huang2021fsdr} also exploits external images, those of ImageNet, for style randomization; the web images are readily available like ImageNet and collected automatically, but more relevant to target application scenarios thanks to the search keyword.

The web-crawled images are often different from synthetic domain images in terms of semantic layout, and could partly contain irrelevant contents due to the ambiguity of search keywords and errors of the search engine.
WEDGE is robust against these issues for the following reasons.
In the first stage of training, the style injection module exploits only styles of the web images while disregarding their contents.
In the second stage, irrelevant parts of an image tend to be ignored in pseudo segmentation labels 
due to their unreliable class predictions (\ie, low confidence).

\begin{figure}[!t]
\begin{center}
\includegraphics[width= \linewidth]{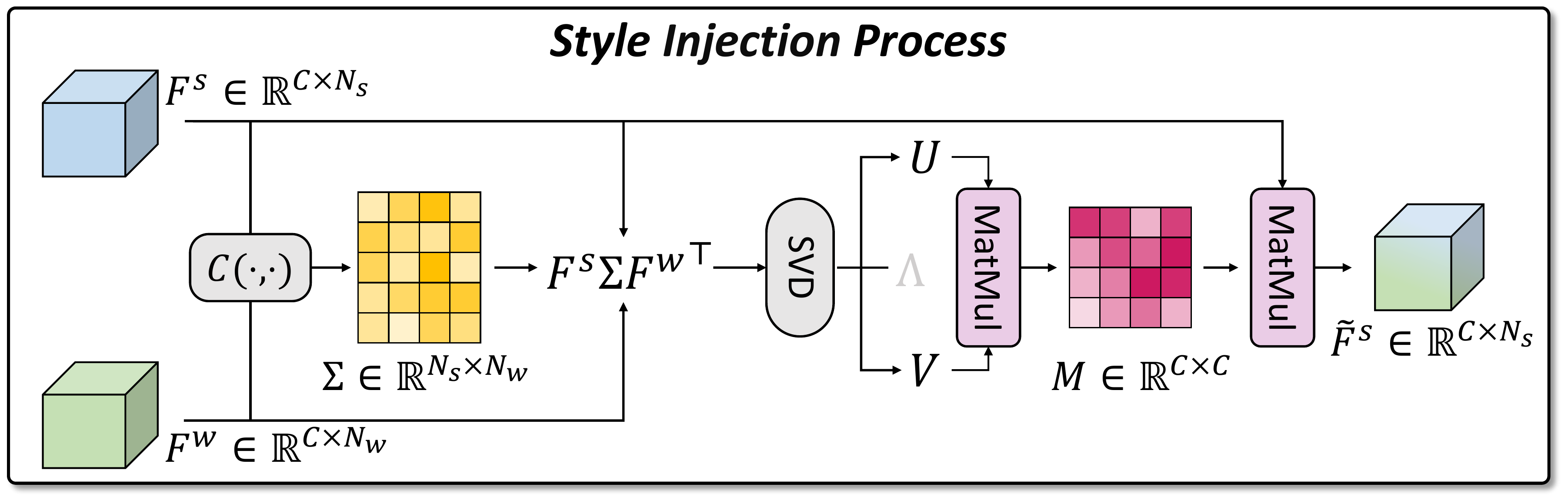}
\end{center}
\vspace{-5mm}
\caption{
Our style injection process.
We first calculate the cross correlation matrix $\Sigma$ between features of source and web images by the cosine similarity $C(\cdot, \cdot)$.
The optimal feature projection matrix for style injection  is computed by $M=UV^\top$, where $U$ and $V$ are obtained by applying SVD to $F^{s}\mathbf{\Sigma}{F^{w}}^{\top}$.
The source feature map is then stylized by applying $M$, and the result $F^{s} M^\top$ is fed into the next convoulution block.
} 
\label{fig:styinj}
\vspace{-4mm}
\end{figure}

\subsection{Stage 1: Learning with Style Injection (SI)}
\label{sec:style_inject}

We propose a content-aware style injection method which transfers styles between semantically similar features of synthetic and web-crawled images.
This method can inject diverse styles while better preserving the content of an image than conventional methods relying on global feature statistics (\eg, AdaIN~\cite{AdaIN} or Gram matrix approximation~\cite{gatys2016image}).

At each iteration of training the segmentation network, a synthetic image $I^s$ is coupled with a randomly sample web image $I^w$.
Let $F^{d,l}\in \mathbb{R}^{H_{d} \times W_{d} \times C}$ be the feature map of $I^d$ from the $l^\textrm{th}$ convolution block of the network where $d\in \{s, w\}$.
First, we compute a cross correlation between the $F^{s,l}$ and $F^{w,l}$ as an affinity matrix $\mathbf{\Sigma}^{l}\in \mathbb{R}^{N_{s}\times N_{w}}$, where 
$N_{d}=H_{d}W_{d}$ ($d\in \{s, w\}$):
\begin{align}
    \mathbf{\Sigma}^{l}_{i,j} = \frac{F^{s,l}_{i}{F^{w,l}_{j}}^{\top}}{\|F^{s,l}_{i}\|\|F^{w,l}_{j}\|}.
\label{eq:similarity}
\end{align}
Then we find the projection matrix $M$ that minimizes the distance between features of the projected feature map $F^{s,l}M^{\top}$ and web-crawled image feature map $F^{w,l}$ weighted by the affinity matrix $\mathbf{\Sigma}^{l}$;
the role of the projection matrix $M$ is to project a synthetic feature onto a subspace of the semantically similar web-crawled features.
The objective function is formulated by
\begin{align}
    \min_{M}J(M) =  \frac{1}{N_{\mathbf{\Sigma}}}\sum^{N_{s}}_{i=1}\sum^{N_{w}}_{j=1} \mathbf{\Sigma}_{ij}^{l} \|F^{s,l}M^{\top}-F_{j}^{w,l}\|,
\label{eq:obj_fun}
\end{align}
where $N_{\mathbf{\Sigma}}$ is sum of all elements of $\mathbf{\Sigma}^{l}$.
The closed form solution of Eq.~\eqref{eq:obj_fun} is given by Huo~\etal~\cite{huo2021manifold} as
\begin{align}
    M = UV^{\top},
\end{align}
where $U$ and $V$ are derived from singular value decomposition of $F^{s,l}\mathbf{\Sigma}^{l}{F^{w,l}}^{\top}$, \ie, $F^{s,l}\mathbf{\Sigma}^{l}{F^{w,l}}^{\top} = U\Lambda{V}^{\top}$.
The synthetic feature map is projected by the estimated $M$, and the result $F^{s,l}M^{\top}$ is fed into the ${l+1}^\textrm{th}$ convolutional block;
\Fig{styinj} illustrates this process.

The overall pipeline of our style injection method follows that of the manifold alignment based style transfer (MAST)~\cite{huo2021manifold}.
However, unlike MAST that computes a discrete affinity matrix using $k$ nearest neighbor assignment, our method uses cosine similarity to compute the continuous affinity matrix in~\Eq{similarity}.
It considers similarity of all features to produce a content-aware projection matrix and does not require hyperparameter $k$ nor $\mathcal{O}(n^2)$ time complexity to assign the nearest neighbors.
Our style injection process is designed as non-parametric, which enables effective and low-cost feature adjustment.

The style injection is applied to multiple convolution blocks of the network, in particular lower blocks since features of deeper layers are known to be less sensitive to style variations. 
More details for implementation can be found in Sec.~\ref{sec:setting}.
Injecting styles of real web images to synthetic training images enlarges the training dataset by a multiple of the number of the web images, which is tremendous regarding the size of the training dataset, as well as making them look diverse and realistic in feature spaces.
In addition, there are several advantages of the content-aware style injection for domain generalization of semantic segmentation over the conventional approaches~\cite{gatys2016image,AdaIN}.
First, it enables style injection between semantically similar regions of web-crawled and synthetic images, which is more natural and effective for semantic segmentation.
Second, since different styles are injected to different semantic regions on an image, it helps keep boundaries between the semantically different regions in style-injected features.
We empirically verify the superiority of our method over other potential style injection candidates in~\Sec{ablation}.

Finally, the network is trained by the pixel-wise cross-entropy loss with the segmentation label of the synthetic image $I^s$.
Let $P^s$ and $Y^s$ denote the segmentation prediction and the groundtruth label of $I^s$, respectively.
The loss is then formulated as
\begin{equation}
    \mathcal{L}_{\textrm{seg}}(P^s, Y^s) = - \frac{1}{N}\sum_{i=1}^H\sum_{j=1}^W\sum_{k=1}^C Y_{ijk}^s \log P_{ijk}^s, \label{eq:loss_pixel_crs_ent}
\end{equation}
where $N=H\times W$.
Although this loss is applied only to the synthetic domain, its gradients with respect to parameters will act as if the network takes real domain images as input thanks to the style injection. 
Note that, in this stage, $I^w$ is used only as a style reference.

\subsection{Stage 2:~Learning Using Pseudo Labels (PL)}
\label{sec:pseudo_label}

Once the first stage is completed, the learned model can be used to generate pseudo labels of the web images.
The pseudo labels allow us to exploit the web images for supervised learning of the segmentation network, which further enhances the generalization capability of the model by learning it directly on a variety of real-world images.

Let $P^w \in \mathbb{R}^{H \times W \times C}$ be the segmentation prediction of the network given $I^w$ as input. 
The pseudo segmentation label of $I^w$, denoted by $\widetilde{Y}^w \in \{0,1\}^{H \times W \times C}$, is obtained by choosing pixels with highly reliable predictions and labeling them with the classes of maximum scores:
\iftrue
\begin{equation} 
    \widetilde{Y}^w_{ijc} = 
    \begin{cases}
            1,  & \textrm{ if } c = \underset{k}{\operatorname{argmax}} \ P^w_{ijk}
                \ \textrm{ and } \ h(P^w_{ij}) < \tau \\
            0, & \textrm{ otherwise}
    \end{cases},
\label{eq:pseudo}
\end{equation}
where $P^w_{ij} \in \mathbb{R}^C$ denotes the class probability distribution of the pixel $(i,j)$, $h(\cdot)$ indicates the entropy, and $\tau$ is a hyperparameter. 
Note that we regard the prediction $P^w_{ij}$ unreliable when its entropy is high, \ie, $h(P^w_{ij}) \geq \tau$; in this case, the pixel is assigned no label and ignored during training.
Fig.~\ref{fig:pseudo_labels_qual} presents examples of the pseudo labels.
\fi

The second stage of training utilizes both of the synthetic and the web images for supervised learning.
It is basically the same with the first stage including the style injection, except that the segmentation loss is now applied to $P^w$ as well as $P^s$.
The total loss for the second stage is thus given by a linear combination of two segmentation losses:
\begin{equation}
    \mathcal{L}(P^s, Y^s, P^w, \widetilde{Y}^w) = \mathcal{L}_\textrm{seg}(P^s, Y^s) + \mathcal{L}_\textrm{seg}(P^w, \widetilde{Y}^w),
\label{eq:pseudo_loss}
\end{equation}
where $\mathcal{L}_\textrm{seg}$ is the cross-entropy loss as given in Eq.~\eqref{eq:loss_pixel_crs_ent}.

\vspace{-1mm}
\section{Experiments}
\label{sec:experiment}

In this section, we first present experimental settings in detail, then demonstrate the effectiveness of WEDGE through extensive results. 
Effectiveness of style injection, pseudo labeling, and other design choices of WEDGE are investigated by ablation studies.

\subsection{Experimental Setting}
\label{sec:setting}

\noindent \textbf{Source datasets.}
As a synthetic source domain for training, we use either the GTA5~\cite{gta5} or the SYNTHIA~\cite{synthia} datasets.
GTA5 consists of 24,966 images and shares the same set of 19 semantic classes with the real test datasets. 
Note that we remove 36 images of smallest file sizes from the dataset since they are non-informative, \eg, blacked-out images.
Meanwhile, SYNTHIA contains 9,400 images, whose annotations cover only 16 classes of the real test datasets. 
Thus, we take only these 16 classes into account when evaluating models trained on SYNTHIA.

\vspace{0.2mm} \noindent \textbf{Test datasets.}
As unseen target domains for evaluation, we choose
the validation splits of Cityscapes~\cite{cityscapes}, BDD100k Segmentation (BDDS)~\cite{BDD100k} and Mapillary~\cite{mapillary}.
Cityscapes and BDDS have 500 and 1,000 validation images, respectively, and they are labeled for the same 19 classes.
2,000 validation images of Mapillary are annotated for 66 classes.
By following the protocol of~\cite{he2020segmentations}, we merge these classes to obtain the same 19 classes of Cityscapes.

\vspace{0.2mm} \noindent\textbf{Web-crawled images.} 
From Flickr, we search for images whose widths are larger than or equal to 760 pixels, and with no copyright reserved (\ie, CC0) for their public use in future work, using the search keyword ``driving + road".
As a result, 4,904 web images in total are collected.
Note that, given these conditions, the crawling process was done automatically, and the retrieved images are used as-is without modification.

\vspace{0.2mm} \noindent \textbf{Networks and their training details.} 
Following previous work~\cite{DRPC}, we adopt DeepLab-v2~\cite{deeplab_v2} with various backbone networks, ResNet50 and ResNet101~\cite{resnet}, as our segmentation networks.
They are first pretrained on ImageNet~\cite{Imagenet},
and then trained with the source dataset and our web images using SGD with momentum of 0.9 and weight decay of $\expnum{5}{4}$.
The initial learning rate is $\expnum{2}{4}$ for the first stage (SI) and $\expnum{1}{4}$ for the second stage (PL). 
$\tau$ in~\Eq{pseudo} is set to $\expnum{5}{2}$ for all experiments.

\vspace{0.2mm} \noindent \textbf{Where to inject styles.} 
Styles of web images are injected into the feature maps output by the $1^\textrm{th}$ and $2^\textrm{nd}$ residual blocks for ResNet101 and ResNet50.
The impact of injection points on performance is analyzed in the accompanying video.

\begin{table}[t!]
\caption{
Quantitative results in mIoU of domain generalization from (G)TA5 to (C)ityscapes, (B)DDS, and (M)apillary.
}
\vspace{-3mm}
\centering
\scalebox{0.97}{
\begin{tabular}{lc|ccc}
\toprule
\multicolumn{1}{l|}{Methods}  & \multicolumn{1}{c|}{~Backbone~~}  &G~$\rightarrow$~C & G~$\rightarrow$~B    & G~$\rightarrow$~M    \\ \midrule
\multicolumn{1}{l|}{IBN-Net~\cite{pan2018two}}  & ResNet50  & 29.6 & - & -   \\ 
\multicolumn{1}{l|}{ASG~\cite{chen2020automated}} & ResNet50  & 31.9 & - & -   \\ 
\multicolumn{1}{l|}{DRPC~\cite{DRPC}} & ResNet50 & {37.4} & {32.1} & {34.1}\\ 
\multicolumn{1}{l|}{RobustNet~\cite{choi2021robustnet}} & ResNet50 & {36.6} & {35.2} & {40.3}\\ 
\multicolumn{1}{l|}{\ccol WEDGE (Ours)~} & \ccol ResNet50 & \ccol \textbf{38.4} & \ccol \textbf{37.0} & \ccol \textbf{44.8} \\\midrule
\multicolumn{1}{l|}{DRPC~\cite{DRPC}} & ResNet101 & {42.5} & {38.7} & {38.1} \\ 
\multicolumn{1}{l|}{FSDR~\cite{huang2021fsdr}} & ResNet101 & 44.8 & 39.7 & 40.9 \\ 
\multicolumn{1}{l|}{PinMemory~\cite{kim2022pin}} & ResNet101 & 44.9 & 39.7 & 41.3 \\
\multicolumn{1}{l|}{\ccol WEDGE (Ours)~} & \ccol ResNet101 & \ccol \textbf{45.2} & \ccol \textbf{41.1} & \ccol \textbf{48.1} \\
\bottomrule
\end{tabular}}
\vspace{-0.5mm}
\label{tab:comp_dg_gta}
\end{table}

\begin{table}[t!]
\caption{
Quantitative results in mIoU of domain generalization from (S)YNTHIA to (C)ityscapes, (B)DDS, and (M)apillary.
}
\vspace{-3mm}
\centering
\scalebox{0.95}{
\begin{tabular}{lc|ccc}
\toprule
\multicolumn{1}{l|}{Methods}  & \multicolumn{1}{c|}{~Backbone~~}  &S~$\rightarrow$~C & S~$\rightarrow$~B    & S~$\rightarrow$~M    \\ \midrule
\multicolumn{1}{l|}{\multirow{1}{*}{DRPC~\cite{DRPC}}} & ResNet50  & 35.7 & 31.5 & 32.7\\
\multicolumn{1}{l|}{\multirow{1}{*}{\ccol WEDGE (Ours)~}} & \ccol ResNet50 & \ccol \textbf{36.1}  & \ccol \textbf{32.5} & \ccol \textbf{37.2} \\ \midrule
\multicolumn{1}{l|}{\multirow{1}{*}{DRPC~\cite{DRPC}}} & ResNet101 & 37.6 & 34.3 & 34.1 \\ 
\multicolumn{1}{l|}{\multirow{1}{*}{FSDR~\cite{huang2021fsdr}}} & ResNet101 & 40.8 & 37.4 & 39.6 \\ 
\multicolumn{1}{l|}{\multirow{1}{*}{\ccol WEDGE (Ours)~}} & \ccol ResNet101 & \ccol \textbf{40.9} & \ccol \textbf{38.1} & \ccol \textbf{43.1} \\
\bottomrule
\end{tabular}
}
\label{tab:comp_dg_syn}
\vspace{-4mm}
\end{table}

\subsection{Comparisons with the State of the Art}
\label{sec:comp_dg}
WEDGE~is compared with existing domain generalization techniques, IBN-Net~\cite{pan2018two}, AGS~\cite{chen2020automated}, DRPC~\cite{DRPC}, RobustNet~\cite{choi2021robustnet}, FSDR~\cite{huang2021fsdr} and PinMemory~\cite{kim2022pin},
using two source domains \{(G)TA5, (S)YNTHIA\}, three test domains \{(C)ityscapes, (B)DDS, (M)apillary\}, and two different backbone networks \{ResNet50, ResNet101\}.
As summarized in Table~\ref{tab:comp_dg_gta} and~\ref{tab:comp_dg_syn}, WEDGE clearly outperforms all the previous arts in all the 12 experiments.

\subsection{In-depth Analysis on WEDGE}
\label{sec:ablation}

\vspace{0.5mm} \noindent \textbf{Detailed performance analysis.}
To investigate the contribution of each training stage in WEDGE, we measure its performance at each stage for all experiments we have conducted so far.
The results in Table~\ref{tab:dgweb} show that 
the first stage using style injection most contributes to the performance in most experiments, which demonstrates the effectiveness of using web-crawled images and our style injection module for domain generalization.
This achievement is remarkable, especially when considering that web images could be erroneous or irrelevant to the test domains.
Thanks to our style injection modules, WEDGE exploits diverse and realistic styles of web images while disregarding their contents that may be irrelevant.
The second stage also leads to non-trivial performance improvement, particularly in the generalization from SYNTHIA to BDDS, which imply the semantics or layouts of pseudo labels on SYNTHIA is more similar to those of BDDS than the other datasets.

\begin{table}[t!]
\centering
\caption{
Performance of WEDGE for domain generalization from (G)TA5 and (S)YNTHIA to (C)ityscapes, (B)DDS, and (M)apillary.}
\vspace{-3mm}
\scalebox{0.85}{
\begin{tabular}{c|ccc|ccc}
\toprule
     & \multicolumn{3}{c|}{ResNet50} & \multicolumn{3}{c}{ResNet101} \\ \midrule
     & \begin{tabular}[c]{@{}c@{}}\textit{~Src.~~}\\ \textit{~~only~~~}\end{tabular} & \begin{tabular}[c]{@{}c@{}}SI\\ \footnotesize{(Stage 1)}\end{tabular} & \begin{tabular}[c]{@{}c@{}}PL\\ \footnotesize{(Stage 2)}\end{tabular} & \begin{tabular}[c]{@{}c@{}}\textit{~Src.~~}\\ \textit{~~only~~~}\end{tabular}  & \begin{tabular}[c]{@{}c@{}}SI\\ \footnotesize{(Stage 1)}\end{tabular}  & \begin{tabular}[c]{@{}c@{}}PL\\ \footnotesize{(Stage 2)}\end{tabular} \\ \midrule
G~$\rightarrow$~C  & 28.29 & 36.25 & 38.36 & 34.28 & 43.55 & 45.18 \\
G~$\rightarrow$~B  & 29.16 & 36.30 & 37.00 & 32.96 & 40.35 & 41.06 \\
G~$\rightarrow$~M  & 40.46 & 42.75 & 44.82 & 41.31 & 47.30 & 48.06 \\ 
G$_\textrm{avg}$   & 32.64 & 38.43 & 40.06 & 36.18 & 43.73 & 44.77 \\ 
\midrule
\midrule
S~$\rightarrow$~C  & 27.06 & 35.28 & 36.09 & 29.96 & 38.22 & 40.94 \\
S~$\rightarrow$~B  & 23.96 & 28.62 & 32.51 & 24.28 & 30.74 & 38.07 \\
S~$\rightarrow$~M  & 31.67 & 36.49 & 37.18 & 36.19 & 38.61 & 43.10 \\
S$_\textrm{avg}$   & 27.56 & 33.46 & 35.26 & 30.14 & 35.86 & 40.70 \\ 
\bottomrule
\end{tabular}
}
\label{tab:dgweb}
\vspace{-0.5mm}
\end{table}

\begin{figure*}[t!]
\begin{center}
\includegraphics[width=0.98\linewidth]{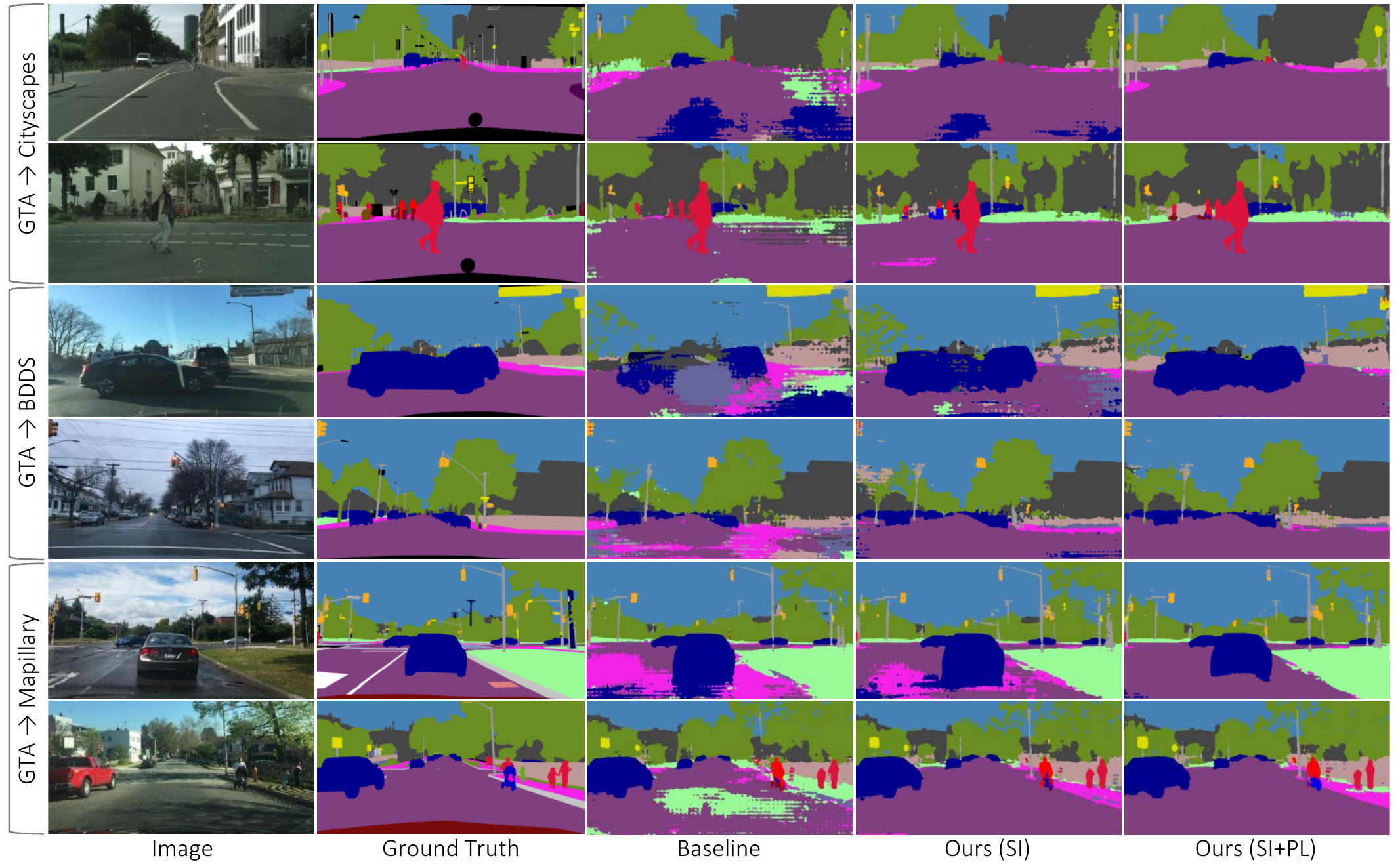}
\end{center}
\vspace{-3mm}
\caption{
Qualitative results of WEDGE and its baseline using ResNet101 backbone and trained on the GTA5 dataset.
} 
\vspace{-1mm}
\label{fig:qual}
\end{figure*}

\vspace{0.5mm} \noindent \textbf{Qualitative results.}
The results in Fig.~\ref{fig:qual} show that the first stage of WEDGE recovers most of the ill-classified pixels, and even finds out objects that are missing in the baseline results.
Also, its second stage further improves the segmentation quality by correcting dotted errors and capturing fine details of object shapes.

\vspace{0.5mm} \noindent \textbf{Comparison of style injection methods.}
We compare our style injection method with other potential candidates based on existing style transfer techniques~\cite{AdaIN,huo2021manifold} to demonstrate its advantages. 
Note that these techniques are also used for injecting styles of web images on-the-fly within the same framework. 
As summarized in~\Tbl{sty_inj_analy}, while using AdaIN~\cite{AdaIN} and MAST~\cite{huo2021manifold} also improves performance, our method achieves the best in both SI and PL stages except for the GTA to BDDS case in the SI stage.
Moreover, our method is more efficient than MAST since it does not need $k$ nearest neighbor search, whose time complexity is $\mathcal{O}(n^2)$, that is required for MAST.

\begin{table}[!t]
\caption{Domain generalization performance of WEDGE with each variant of style injection methods and ours. 
}
\vspace{-3mm}
\centering
\scalebox{0.85}{
\begin{tabular}{l|ccc|ccc}
\toprule
\multirow{2}{*}{SI methods} &  \multicolumn{3}{c|}{SI (Stage 1)} & \multicolumn{3}{c}{PL (Stage 2)} \\ 
 & G$\rightarrow$C & G$\rightarrow$B & G$\rightarrow$M & G$\rightarrow$C & G$\rightarrow$B & G$\rightarrow$M \\ \midrule
None (source only)              & 34.28 & 32.96  & 41.31  & - &  - & - \\
WEDGE$+$AdaIN~\cite{AdaIN} &  40.54 &  {40.78} & 46.33 & 40.58  & 39.74  & 46.92 \\
WEDGE$+$MAST~\cite{huo2021manifold} &  41.69 & 40.05 & 46.01 & 43.17 & 40.34 & 46.28 \\
\ccol WEDGE (Ours) &   \ccol {43.55} &  \ccol {40.35} &  \ccol {47.30} &  \ccol {45.18} &  \ccol {41.06} &  \ccol {48.06}
\\ \bottomrule
\end{tabular}
}
\vspace{-4mm}
\label{tab:sty_inj_analy}
\end{table}

\vspace{0.5mm} \noindent \textbf{Impact of the number of web images.} 
We investigate the impact of the number of web images by evaluating performance of a segmentation model trained by WEDGE with different numbers of web images.
\begin{figure}[!t]
\vspace{-2mm}
\centering
\includegraphics[width=0.9\linewidth]{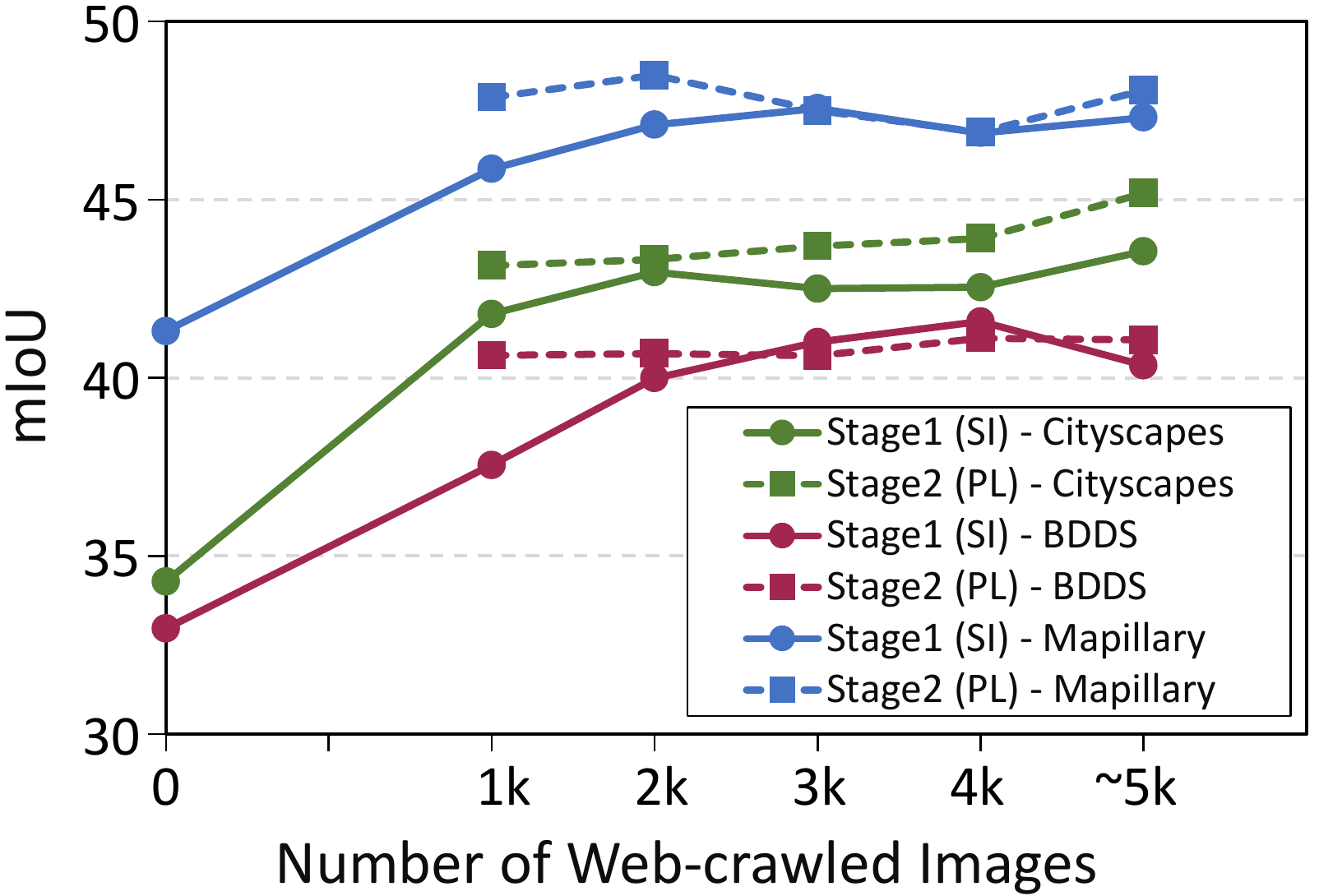}
\vspace{-2mm}
\caption{
Domain generalization performance of WEDGE versus the number of web images. 
In all experiments ResNet101 is used as backbone.
}
\label{fig:abl_web}
\vspace{-7mm}
\end{figure}

In Fig.~\ref{fig:abl_web}, these models are compared in terms of segmentation quality on the three target datasets.
As shown in the figure, the generalization capability of the model can be substantially improved by using only 1,000 web images, while using the whole web dataset further improves performance.
To be specific, when using 1,000 web images, the average mIoU over the 3 test datasets is 42.4\%, lacking only 2.4\% compared to the average performance of our final model.
The results also indicate that WEDGE consistently enhances the generalization performance when increasing the number of web-crawled images.

\vspace{1mm} \noindent \textbf{Impact of using task-relevant web images.} 
The contribution of our crawling strategy is demonstrated by comparing WEDGE with its variants relying on other types of style references instead of the task-relevant web images. 
Specifically, we utilize images sampled from the ImageNet dataset and web images crawled by the search keyword ``indoor'', both of which are irrelevant to the target task.
Also, the number of style references is set to 5,000 for fair comparisons to WEDGE.
Note that since these images are totally irrelevant to the target task, they are not suitable for pseudo labeling thus are used only for style injection.
As shown in \Tbl{abl_query}, our method using task-relevant web images (\ie, ``driving+road'') clearly outperforms the others. 
Using the real yet irrelevant images improves performance, suggesting the robustness of our method,
but the results are still inferior to those of our method, meaning that our crawling strategy is useful and using relevant images matters.

\begin{table}[!t]
\centering
\caption{
Domain generalization performance of WEDGE with different types of style reference.
}
\vspace{-3mm}
\scalebox{1}{
\begin{tabular}{l|ccc}
\toprule
Style reference         & ~G~$\rightarrow$~C~ & G~$\rightarrow$~B & ~G~$\rightarrow$~M~ \\ \midrule
None (source only)              & 34.28 & 32.96  & 41.31 \\
ImageNet                        & 42.42 & 38.24  & 47.37 \\
Web images: ``indoor''          & 42.28 & 40.47  & 47.69 \\
Web images: ``driving+road''~    & 45.18 & 41.06 & 48.06 \\ 
\bottomrule
\end{tabular}
}
\vspace{-5mm}
\label{tab:abl_query}
\end{table}

\section{Conclusion}
\label{sec:conclusion}

We have introduced WEDGE, the first web-image assisted domain generalization scheme for learning semantic segmentation. 
It explores and exploits web images that depict large diversity of real world scenes, which potentially cover latent test domains and thus help improve generalization capability of trained models.
It utilizes the web-crawled images in two effective ways,
namely style injection and pseudo labeling, which lead to consistent performance improvement on various test domains.
WEDGE clearly outperformed existing domain generalization techniques in all experiments.
Also, extensive ablation studies demonstrated that WEDGE is able to utilize noisy and irrelevant web-crawled images reliably and is not sensitive to their number in training.

\vspace{3mm}
{\small
\noindent \textbf{Acknowledgement.} 
This work was supported by Samsung Research Funding \& Incubation Center of Samsung Electronics under Project Number SRFC-IT1801-52.
This work was done while Namyup Kim was working as an intern at Microsoft Research Asia.
}

\pagebreak

\noindent \textbf{\huge{Appendix}}
\vspace{5mm}

This material presents implementation details and experimental results omitted from the main paper due to the space limit. First, \Sec{ben_styinj} demonstrates advantages of the feature-level style injection compared to the image-level style transfer.
\Sec{tau} examines how much sensitive WEDGE is to the hyper-parameter~$\tau$ and \Sec{styinj} investigates the impact of style injection points by an ablation study.  
\Sec{morekeywords} discusses using domain-specific web images and \Sec{webimage} provides more qualitative examples of the web-crawled images we use.
Then \Sec{morequal} presents more qualitative results of WEDGE.

\section{Advantages of Feature-level Style Injection}
\label{sec:ben_styinj}
\begin{figure}[t!]
\vspace{12mm}
\begin{center}
\includegraphics[width=\linewidth]{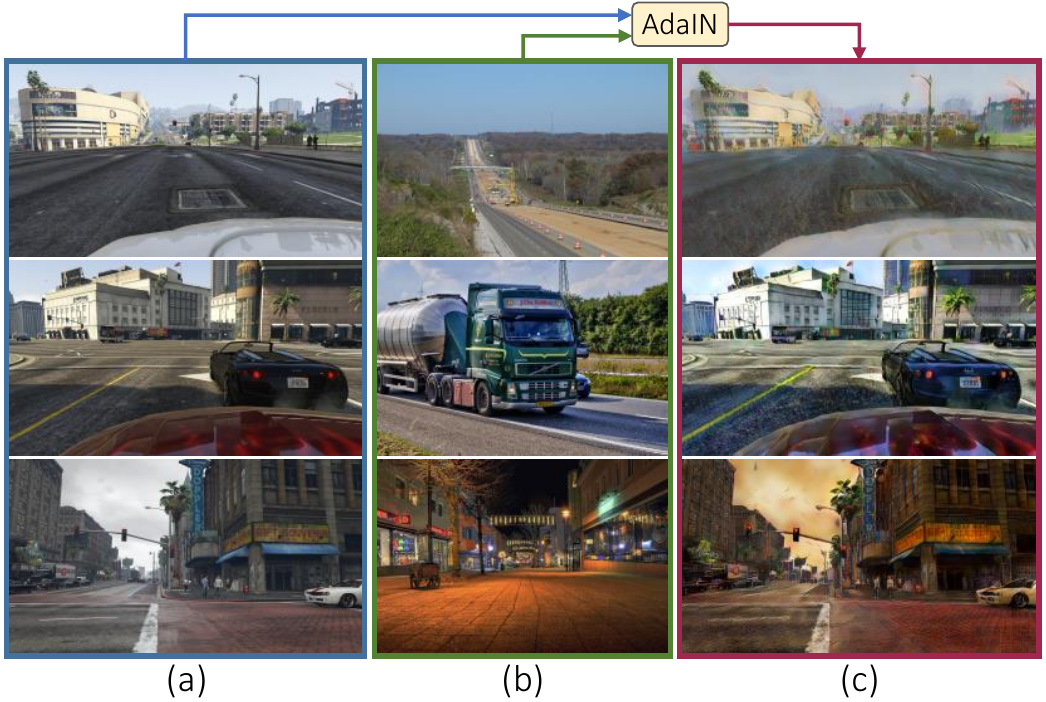}
\end{center}
\vspace{-3mm}
\caption{Examples of GTA5 images stylized by AdaIN. (a) Content images. (b) Style images. (c) Stylized images.}
\label{fig:adain}
\end{figure}

Since WEDGE injects style representations in feature levels, one may wonder its advantages over image-level style transfer.
This section demonstrates the effectiveness of WEDGE, especially its style injection (SI) module, compared to image-level style transfer.
To this end, we adopt AdaIN~\cite{AdaIN}, exploiting feature statistics as style representation like WEDGE.
We generate 100,000 stylized GTA5~\cite{gta5} images by AdaIN using web-crawled images as style references; a few examples are shown in~\Fig{adain}.
We then train a segmentation model on the stylized GTA5 dataset.

As summarized in~\Tbl{ada_styinj}, we compare the model of the 1$^\textrm{st}$ stage of WEDGE (SI only) with the model trained on the stylized GTA5 images generated by AdaIN.
The results show WEDGE using SI outperforms AdaIN on all experimental settings except G$\rightarrow$C and G$\rightarrow$B with VGG16~\cite{resnet}.
Moreover, our feature-level approach has another benefit over the image-level counterpart in terms of efficiency.
AdaIN requires an additional network for style transfer.
On the other hand, SI in WEDGE is non-parametric and adjusting feature statistics of source images by those of web-crawled images, thus demands a much lower computational cost than AdaIN.

\section{Sensitivity to Hyper-parameter $\tau$}
\label{sec:tau}
This section demonstrates the impact of the thresholding parameter s$\tau$ on the quality of pseudo labels in terms of semantic segmentation performance.
Specifically, pseudo segmentation labels are generated using $\tau$, which is a hyper-parameter that filters out unreliable predictions.
To this end, we design multiple variants of our model that are trained from different pseudo segmentation labels generated from various $\tau$.
The pseudo segmentation label of $I^w$, denoted by $\widetilde{Y}^w \in \{0,1\}^{H \times W \times C}$, is obtained by choosing pixels with highly reliable predictions and labeling them with the classes of maximum scores:
\begin{equation} 
    \widetilde{Y}^w_{ijc} = 
    \begin{cases}
            1,  & \textrm{ if } c = \underset{k}{\operatorname{argmax}} \ P^w_{ijk}
                \ \textrm{ and } \ h(P^w_{ij}) < \tau \\
            0, & \textrm{ otherwise}
    \end{cases},
\label{eq:pseudo_supp}
\end{equation}

where $h(\cdot)$ indicates the entropy, $P^w_{ij} \in \mathbb{R}^C$ denotes the class probability distribution of the pixel $(i,j)$, and $\tau$ is a hyper-parameter. 
We sample $\tau$ from \{0.1, 0.05, 0.01, 0.005\}, where $\tau=0.05$ means our model in the main paper.

\begin{figure}[!t]
\begin{center}
\includegraphics[width=\linewidth]{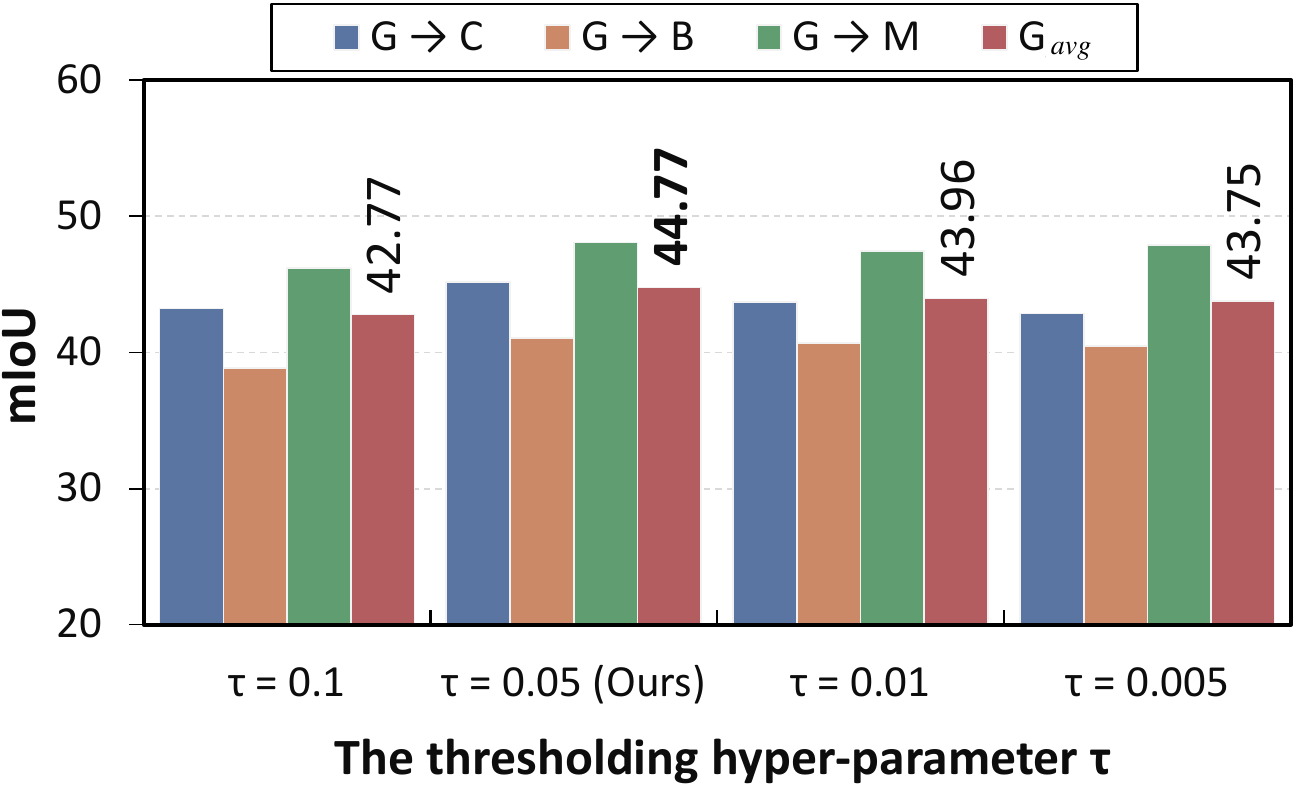}
\end{center}
\vspace{-3mm}
\caption{
Performance of our ResNet101 backbone model trained with different thresholded pseudo labels.
$\textrm{G}_{avg}$ is the average performance of the three test domains. 
The performances across the set of hyper-parameters $\tau$ except 0.1 are marginally different.
}
\label{fig:abl_tau}
\end{figure}

\begin{figure*}[t!]
\begin{center}
\includegraphics[width=0.95 \linewidth]{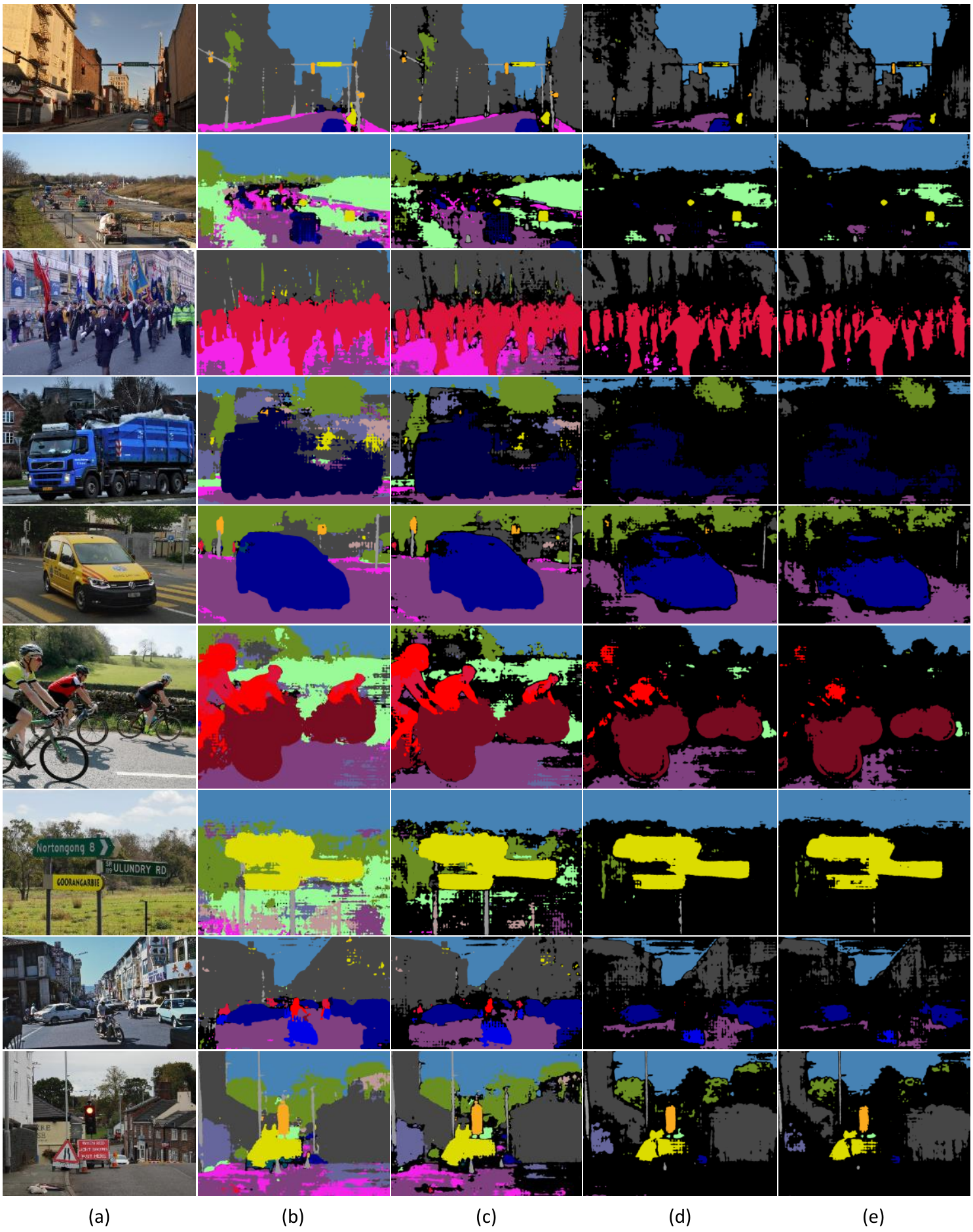}
\end{center}
\vspace{-4mm}
\caption{Examples of the pseudo labels with the different thresholding hyper-parameter $\tau$. (a) Input images. Pseudo labels with (b) $\tau=0.1$. (c) $\tau=0.05$ (Ours). (d) $\tau=0.01$. (e) $\tau=0.005$.}
\label{fig:pseudo_tau}
\vspace{-4mm}
\end{figure*}

As summarized in~\Fig{abl_tau}, the results are marginally different across the variation of the hyper-parameters except 0.1, but the setting we adopt in the paper is slightly better than the others.
Examples of the pseudo labels of web-crawled images are presented in~\Fig{pseudo_tau}, which demonstrates both pros and cons of different threshold values.
With a moderate thresholding (\eg, 0.1), the pseudo labels cover more real texture or parts of an object but have more noisy semantic labels.
With a strict thresholding, on the other hand, the pseudo labels have more accurate semantic information but cover smaller regions of web-crawled images.
The thresholding hyper-parameter we choose is in the middle, and leads to the best performance.

\begin{table}[t!]
\caption{Quantitative results in mIoU and parameters of domain generalization from (G)TA5 to (C)ityscapes , (B)DDS and (M)appillary.
}
\centering
\vspace{-2mm}
\scalebox{0.9}{
\begin{tabular}{cccccc}
\toprule
\multicolumn{1}{c|}{Methods}  & \multicolumn{1}{c|}{Backbone} & \multicolumn{1}{c|}{Params} & G~$\rightarrow$~C & G~$\rightarrow$~B    & G~$\rightarrow$~M    \\ \midrule
\multicolumn{1}{c|}{\multirow{3}{*}{\begin{tabular}[c]{@{}c@{}}Deeplab-v2~\cite{deeplab_v2}\\ +AdaIN\end{tabular}}} & \multicolumn{1}{c|}{VGG16} & \multicolumn{1}{c|}{53.1M} & 35.33 & 34.49 & 40.17\\
\multicolumn{1}{c|}{} & \multicolumn{1}{c|}{ResNet50} & \multicolumn{1}{c|}{48.6M}  & 33.31 &  {34.02} & 38.55\\
\multicolumn{1}{c|}{} & \multicolumn{1}{c|}{ResNet101} & \multicolumn{1}{c|}{67.6M} & 39.41 & 36.20 & 41.50 \\ \midrule
\multicolumn{1}{c|}{\multirow{3}{*}{~~WEDGE (SI)~~}} & \multicolumn{1}{c|}{VGG16} & \multicolumn{1}{c|}{29.6M} &  {35.33} &  {34.48} &  {40.54}    \\
\multicolumn{1}{c|}{}  & \multicolumn{1}{c|}{ResNet50} & \multicolumn{1}{c|}{25.1M} &  {36.25}  & 36.30 &  {42.75} \\
\multicolumn{1}{c|}{}  & \multicolumn{1}{c|}{ResNet101} & \multicolumn{1}{c|}{44.0M} &   {43.55} &  {40.35} &  {47.30} \\
\bottomrule
\end{tabular}
}
\label{tab:ada_styinj}
\end{table}

\section{Details of Style Injection}
\label{sec:styinj}

Style representations of web-crawled images are injected into the feature maps output by $1^\textrm{th}$ and $2^\textrm{nd}$ residual blocks for both ResNet101 and ResNet50 ~\cite{resnet}, and those of $2^\textrm{nd}$ and $3^\textrm{rd}$ blocks for VGG16~\cite{vggnet}.
To verify the effectiveness of our method, this section presents ablation studies with various combinations of injection points. 
We present experimental results with ResNet101 combined with six different combinations in ~\Tbl{styinj_101}.
The results show that semantic segmentation performance is degraded when $4^\textrm{th}$ residual block is included.
We suspect this is because deeper features are known to contain semantic information rather than styles, which makes them inappropriate for style injection.
As a result, using the output feature maps from the \{$1^\textrm{st}$, $2^\textrm{nd}$\} residual blocks turn out to be the most effective combination for ResNet101.
Therefore, we choose our injection points based on these observations when applying style injection to other backbone networks.

\begin{table}[t!]
\vspace{1mm}
\caption{Performance of the models with ResNet101 backbone on the setting from (G)TA5 to (C)ityscapes, (B)DDS and (M)appillary.
}
\vspace{-1mm}
\centering
\scalebox{1}{
\begin{tabular}{cccc|cccc}
\toprule
\multicolumn{4}{c|}{Style injection points} & \multirow{2}{*}{G~$\rightarrow$~C} & \multirow{2}{*}{G~$\rightarrow$~B} & \multirow{2}{*}{G~$\rightarrow$~M} & \multirow{2}{*}{Average} \\
1 & 2 & 3 & 4 &  &  &  &  \\ \midrule
\cmark & \cmark &  &                & 43.55 & 40.35 & 47.30 & 43.73 \\
\cmark & \cmark & \cmark &          & 44.03 & 39.30 & 47.30 & 43.54 \\
\cmark & \cmark & \cmark & \cmark   & 41.01 & 38.49 & 46.46 & 41.99 \\
 & \cmark & \cmark &                & 42.00 & 39.03 & 44.93 & 41.99 \\
 &  & \cmark & \cmark               & 37.92 & 35.02 & 38.47 & 37.20 \\
 & \cmark & \cmark & \cmark         & 38.64 & 35.19 & 41.39 & 38.44 \\ 
\bottomrule
\end{tabular}
}
\label{tab:styinj_101}
\end{table}

\begin{table}[!t]
\centering
\vspace{1mm}
\caption{
Domain generalization performance of WEDGE with different types of style reference.
In all experiments ResNet101 is used as backbone.
}
\vspace{-1mm}
\scalebox{0.92}{
\begin{tabular}{l|cc|cccc}
\toprule
Method & \multicolumn{2}{c|}{Keywords} & G$\rightarrow$C & G$\rightarrow$B & G$\rightarrow$M & Average \\ \midrule
\multirow{4}{*}{WEDGE (SI)} & \multicolumn{2}{c|}{driving + snow} & 42.44 & 39.78 & 45.13 & 42.45 \\
 & \multicolumn{2}{c|}{driving + rain} & 42.87 & 38.13 & 45.95 & 42.32 \\
 & \multicolumn{2}{c|}{driving + fog} & 43.40 & 40.22 & 46.54 & 43.39 \\
 & \multicolumn{2}{c|}{driving + road} & \textbf{43.55} & \textbf{40.35} & \textbf{47.30} & \textbf{43.73}\\ 
 \bottomrule
\end{tabular}
}
\label{tab:table_more_keywords}
\end{table}

\section{Comparison with using domain-specific web images}
\label{sec:morekeywords}
Since our task at hand is domain generalization that assumes arbitrary target domains, we employ the keyword that does not indicate any specific domains.
Nevertheless, we experiment with the keywords ``driving + \{rain, show, fog\}''.
As summarized in~\Tbl{table_more_keywords}, these specific keywords are not as useful as the general one ``driving + road'' in our framework.

\section{Examples of Web-crawled Images}
\begin{figure*}[t!]
\begin{center}
\includegraphics[width=0.95\linewidth]{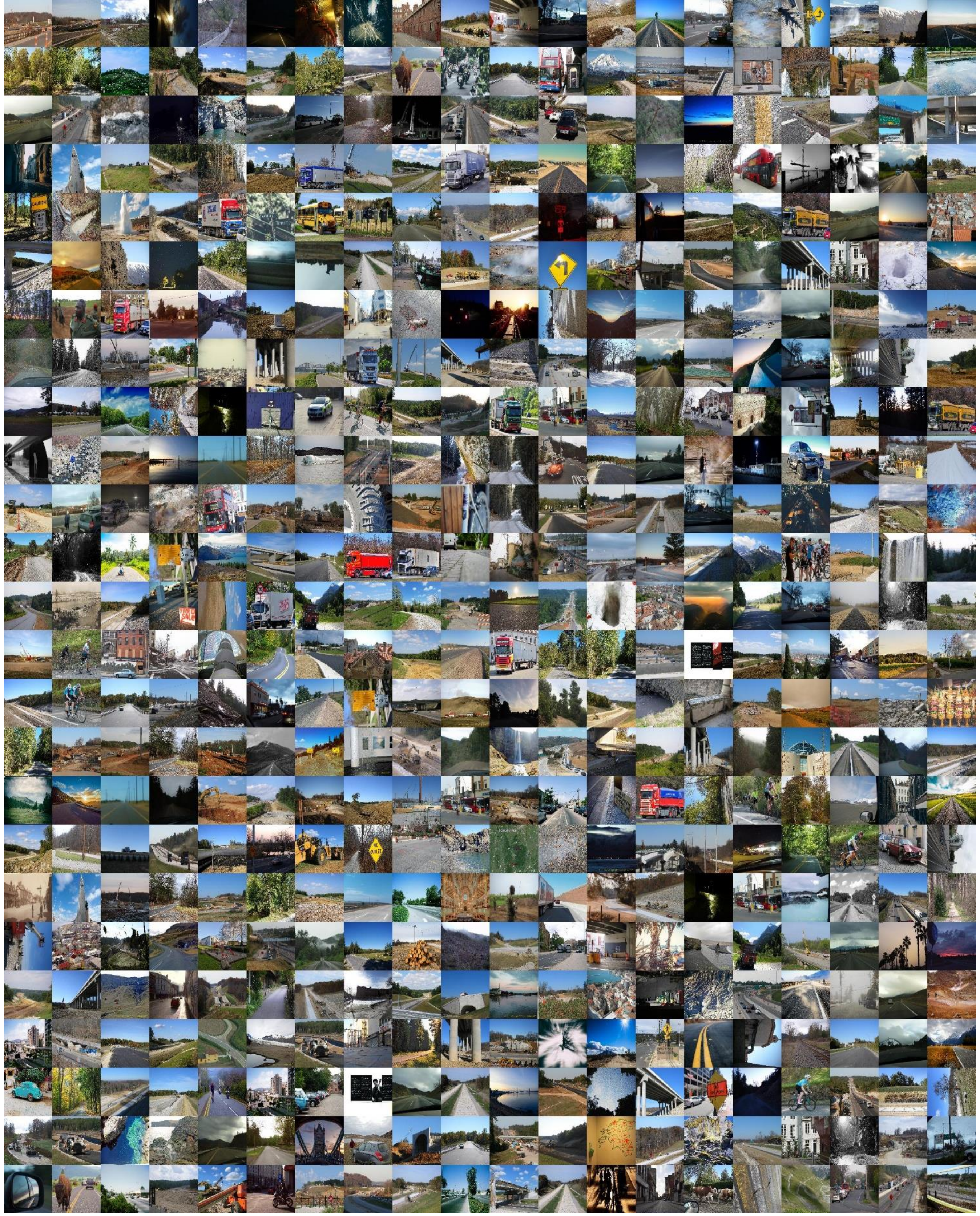}
\end{center}
\vspace{-2mm}
\caption{
500 random samples of the web-crawled images used in WEDGE.
} 
\label{fig:supp_webimage}
\end{figure*}

\vspace{-1mm}
\label{sec:webimage}
This section exhibits a part of our web-crawled dataset.
Qualitative examples of the web-crawled images are presented in Fig.~\ref{fig:supp_webimage},
which demonstrates the diversity of the images in terms of time, geolocation, weather condition, and so on.
Such diversity enables WEDGE to achieve the generalization to latent real domains.
Note that these images often depict entities and semantic layouts that are irrelevant to those of source (and target) domains.
However, they are used as-is with no manual filtering process since the style injection and pseudo labeling of WEDGE offer reliable and effective ways to utilize them.

\section{More Qualitative Results}
\label{sec:morequal}
\begin{figure*}[t!]
\begin{center}
\vspace{15mm}
\includegraphics[width=0.95 \linewidth]{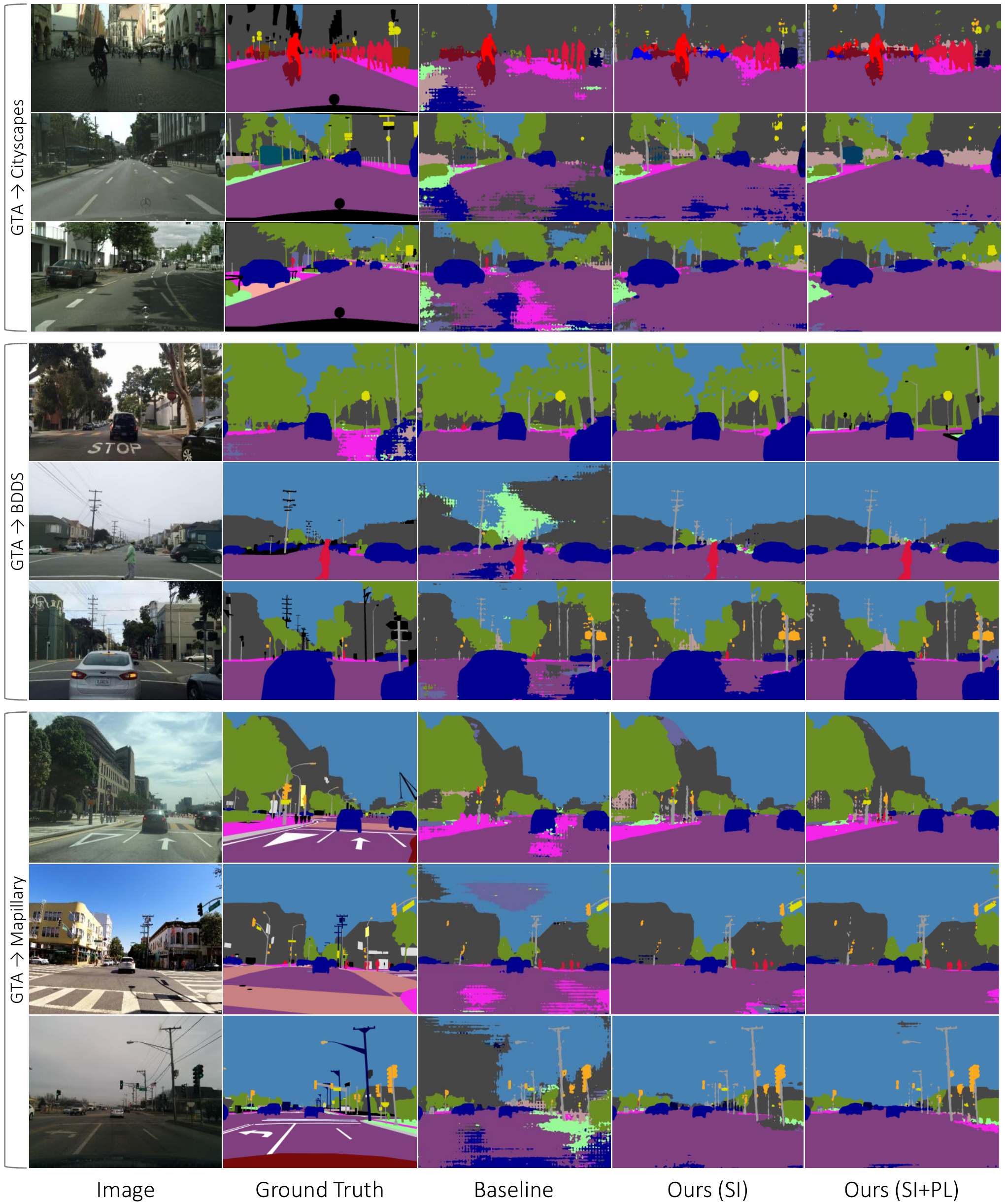}
\end{center}
\vspace{-2mm}
\caption{
Qualitative results of WEDGE and its baseline using ResNet101 backbone and trained on the GTA5 dataset.
} 
\vspace{15mm}
\label{fig:supp_qual1}
\end{figure*}

\begin{figure*}[t!]
\begin{center}
\vspace{15mm}
\includegraphics[width=0.95 \linewidth]{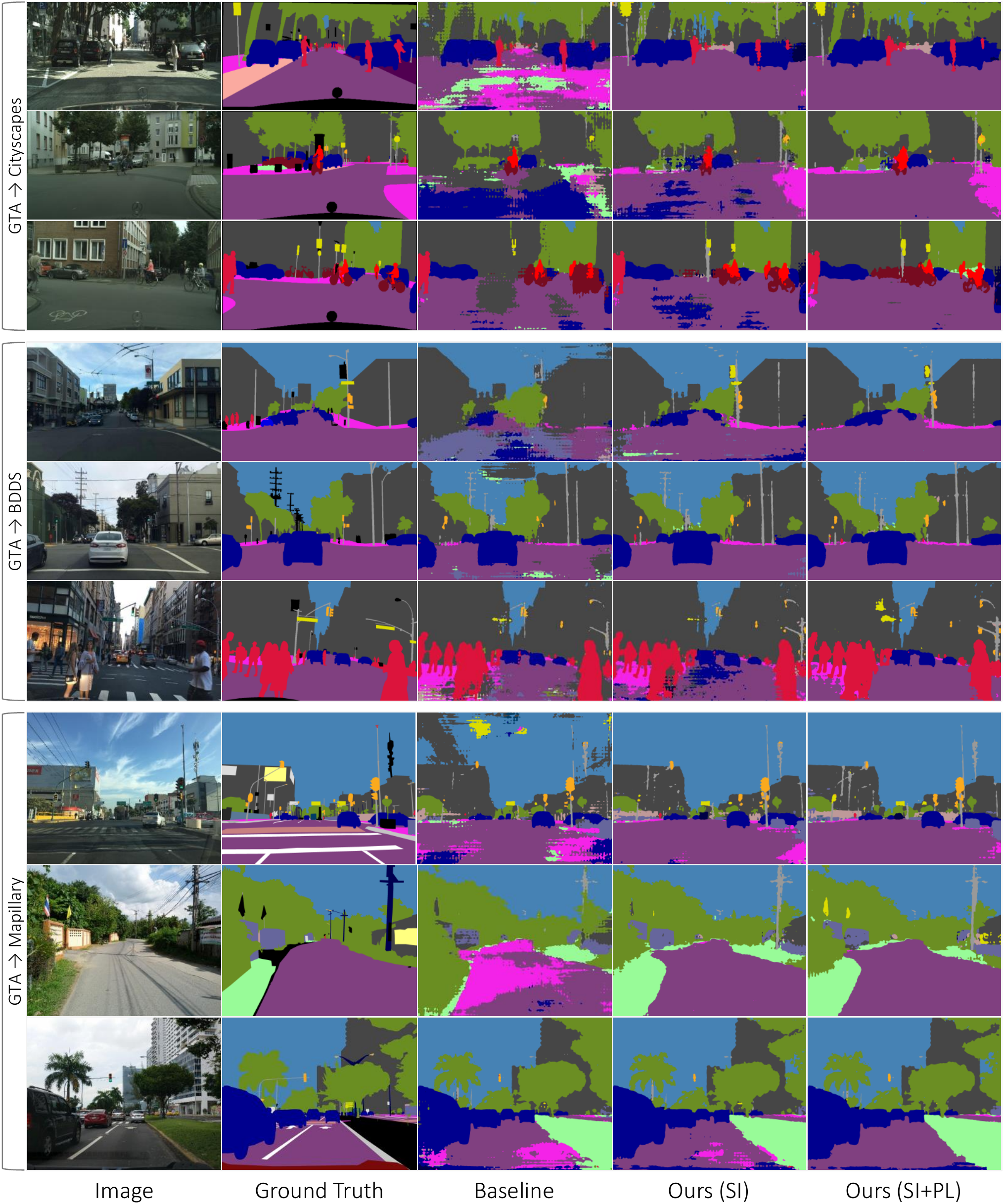}
\end{center}
\vspace{-2mm}
\caption{
Qualitative results of WEDGE and its baseline using ResNet101 backbone and trained on the GTA5 dataset.
} 
\vspace{15mm}
\label{fig:supp_qual2}
\end{figure*}

We present qualitative examples of semantic segmentation results by WEDGE for both of Stage 1 (SI) and Stage 2 (SI+PL) in~\Fig{supp_qual1} and~\Fig{supp_qual2}.
In these figures, the semantic segmentation results are color-coded by following the standard Cityscapes color map~\cite{cityscapes}; the colors associated to the classes are exhibited in~\Tbl{colorcode}.

\begin{table}[!h]
\centering
\scalebox{1.2}{
\begin{tabular}{cccc}
\cellcolor{aero}road & \cellcolor{bicy}sidewalk   & \cellcolor{bird}\brt{building}        & \cellcolor{boat}\brt{wall}        \\
\cellcolor{bottle}fence    & \cellcolor{bus}pole       & \cellcolor{car}traffic light         & \cellcolor{cat}traffic sign        \\
\cellcolor{chair}vegetation     & \cellcolor{cow}terrain       & \cellcolor{table}sky & \cellcolor{dog}person         \\
\cellcolor{horse}rider     & \cellcolor{mbike}\brt{car} & \cellcolor{person}\brt{truck}      & \cellcolor{plant}\brt{bus} \\
\cellcolor{sheep}\brt{train}     & \cellcolor{sofa}\brt{motorbike}      & \cellcolor{train}\brt{bicycle}       &   \\
\end{tabular}
}
\vspace{-1mm}
\caption{The color code of classes on the test datasets.}
\vspace{-4mm}
\label{tab:colorcode}
\end{table}

\clearpage

\clearpage

\pagebreak
{\small
\bibliographystyle{IEEEtran}
\bibliography{bibtax}
}

\end{document}